\documentclass[journal]{IEEEtran}
\usepackage{amsmath,amsfonts}
\usepackage{amssymb}
\usepackage[ruled,vlined]{algorithm2e}
\usepackage{array}
\usepackage{multirow}
\usepackage{booktabs}
\usepackage[caption=false,font=normalsize,labelfont=sf,textfont=sf]{subfig}
\usepackage{textcomp}
\usepackage{stfloats}
\usepackage{url}
\usepackage{verbatim}
\usepackage{cite}
\usepackage[table]{xcolor}
\usepackage{tabularx}
\usepackage{graphicx}
\usepackage{tikz}
\usepackage{paralist}
\usepackage{capt-of}
\usepackage{placeins}

\newcommand{\ie}{i.e.}

\definecolor{myy}{RGB}{126,95,0}
\definecolor{mygray}{gray}{.9}
\definecolor{bblue}{RGB}{30,80,120}
\definecolor{mygray1}{gray}{.7}

\definecolor{ggray}{RGB}{127,127,127}
\definecolor{mygreen}{RGB}{93,174,86}

\definecolor{rblue}{rgb}{0,0.5,1}
\definecolor{hollywoodcerise}{rgb}{0.96, 0.0, 0.63}
\definecolor{lasallegreen}{rgb}{0.03, 0.47, 0.19}
\definecolor{hanpurple}{rgb}{0.32, 0.09, 0.98}
\definecolor{green(pigment)}{rgb}{0.0, 0.65, 0.31}

\usepackage[pagebackref=false,breaklinks=true,colorlinks,bookmarks=false]{hyperref}
\hypersetup{colorlinks,linkcolor={red},citecolor={hanpurple},urlcolor={magenta}} 
\newcommand{\changed}[1]{{\textcolor{black}{#1}}}
\newcommand{\changedv}[1]{{\textcolor{black}{#1}}}

\begin{document}

\title{LFX: Fusion-Form-Flexible Light Field Modeling for Dense Semantic Segmentation and Salient\\Object Detection}
\author{Fei Teng$^{1,*}$, Lingxin Huang$^{1,*}$, Buyin Deng$^{1}$, Kai Luo$^{1}$, Boyuan Zheng$^{1}$, Zheng Fang$^{2}$, Hong Zheng$^{2}$,\\Kunyu Peng$^{3}$, Jiaming Zhang$^{1}$, Yaonan Wang$^{1}$, and Kailun Yang$^{1}$,~\IEEEmembership{Member,~IEEE}%
\thanks{This work was supported in part by the National Natural Science Foundation of China (Grant No. 62388101 and No. 62473139), in part by the Hunan Provincial Research and Development Project (Grant No. 2025QK3019), in part by the State Key Laboratory of Autonomous Intelligent Unmanned Systems (the opening project number ZZKF2025-2-10), and in part by China Mobile Hunan Company Limited and China Mobile Communications Group Co., Ltd., and was conducted under the project ``Research on Reinforcement Learning Algorithm for Quadruped Bionic Locomotion with Emotional Expression''.
\textit{(Corresponding author: Kailun Yang.)}}
\thanks{$^{1}$The authors are with the School of Artificial Intelligence and Robotics and the National Engineering Research Center of Robot Visual Perception and Control Technology, Hunan University, China (email: kailun.yang@hnu.edu.cn).}
\thanks{$^{2}$The authors are with China Mobile Group Hunan Company Ltd., China.}
\thanks{$^{3}$The author is with the Institute for Anthropomatics and Robotics, Karlsruhe Institute of Technology, Germany.}
\thanks{$^{*}$Equal contribution.}
}

\markboth{IEEE Transactions on Circuits and Systems for Video Technology, November~2026}%
{Teng~\MakeLowercase{\textit{et al.}}: LFX}


\let\oldtwocolumn\twocolumn
\renewcommand\twocolumn[1][]{%
    \oldtwocolumn[{#1}{
    \begin{center}
    \vskip-2ex
        \includegraphics[width=0.95\textwidth]{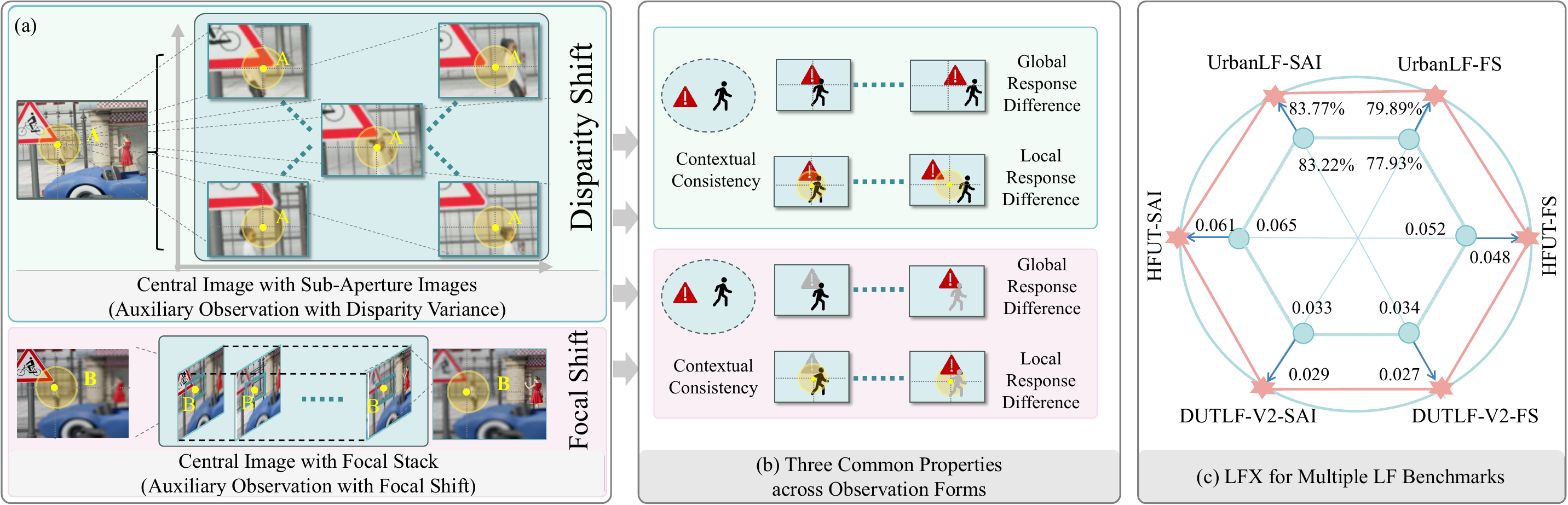}
        \captionof{figure}{\small\changedv{Overview of LFX for form-flexible LF
perception. (a) Two representative LF observation forms relative to the
central image: S.Aperture images with viewpoint-induced disparity shifts and
FS with focal-plane-dependent focus changes. (b) Despite their different
formation processes, both forms exhibit contextual consistency, global
response difference, and local response difference. (c) The same LFX design
is instantiated and trained separately under the two LF forms and achieves the best results on DUTLF-V2, HFUT, and UrbanLF-real.}}
        \label{fig:Fig12}
    \end{center}
    }]
}
\maketitle

\begin{abstract}

Two distinct light field (LF) observation forms, namely sub-aperture images (S.Aperture images) and focal stack (FS), correspond to two fusion settings: the central image with S.Aperture images (CS) and the central image with FS (CF). Existing LF perception methods are typically form-specific, while a modeling design that can be separately instantiated under either setting remains underexplored. To address this gap, we present LFX, a form-flexible LF perception framework. Here, ``X'' represents the LF observation form, indicating that the one modeling design can be separately instantiated under two fusion settings (CS and CF). Specifically, we propose Field-of-Parallax Angular Subspace Modeling (FoP-ASM), which maps the feature difference between each auxiliary observation and the central image to an observation-specific angular marker. This avoids pairwise modeling among auxiliary observations and form-specific geometric assumptions. Meanwhile, a shared manifold subspace and the associated regularization losses constrain the angular residuals within a common feature space, promoting semantic consistency across auxiliary observations. Extensive evaluations across three LF datasets show that LFX achieves state-of-the-art results across distinct LF forms. Under the S.Aperture images and FS settings, respectively, LFX reduces MAE by $12.12\%/20.59\%$ on DUTLF-V2 and by $6.15\%/7.69\%$ on HFUT, while achieving the best performance on UrbanLF-real. The source code will be made publicly available at \href{https://github.com/FeiT-FeiTeng/LFX}{LFX}.

\end{abstract}


\section{Introduction}
\IEEEPARstart{L}{ight} Field (LF) cameras have attracted increasing attention in a wide range of applications, including autonomous driving~\cite{cong2024multimodal,cmnext}, 3D reconstruction~\cite{3dr1,3dr2,3dr3,chen2024pixel}, and virtual reality~\cite{ar1,ar2,ar3}. 
Unlike conventional pinhole cameras that capture a single viewpoint in one shot, LF cameras record multiple perspectives within a single exposure, introducing angular information that elevates planar perception to spatially aware understanding without requiring additional hardware~\cite{chen2024view,lfimaging,lfimaging2}.

To formally characterize such multi-perspective sensing, LF observations are typically represented as a \changed{four-dimensional function $L(u,v,x,y)$}, where angular coordinates $(u,v)$ augment planar coordinates $(x,y)$. Based on this angular-spatial parameterization, an LF capture can be post-processed into two typical observation forms, distinguished by the pairwise relationships among their auxiliary observations~\cite{ihrke2016principles,lightreview,lightreview2,PKU,wang2023light}, as illustrated in Fig.~\ref{fig:Fig12}(a).
(1) The \emph{disparity-centric} form~\cite{jia2024prompt,ye2023lfienet} uses sub-aperture images (S.Aperture images) as auxiliary observations to encode angular structure through cross-view pixel disparities. This form leads to the fusion of the central image with S.Aperture images (CS).
(2) The \emph{refocus-centric} form~\cite{liu2023lftransnet,focalstack1,focalstack2,liu2024lfsamba} reparameterizes angular variations into multiple spatially aligned refocused observations, collectively forming a focal stack (FS). This form leads to the fusion of the central image with FS (CF). 

\begin{figure}[!t]
    \centering
    \includegraphics[width=0.96\linewidth]{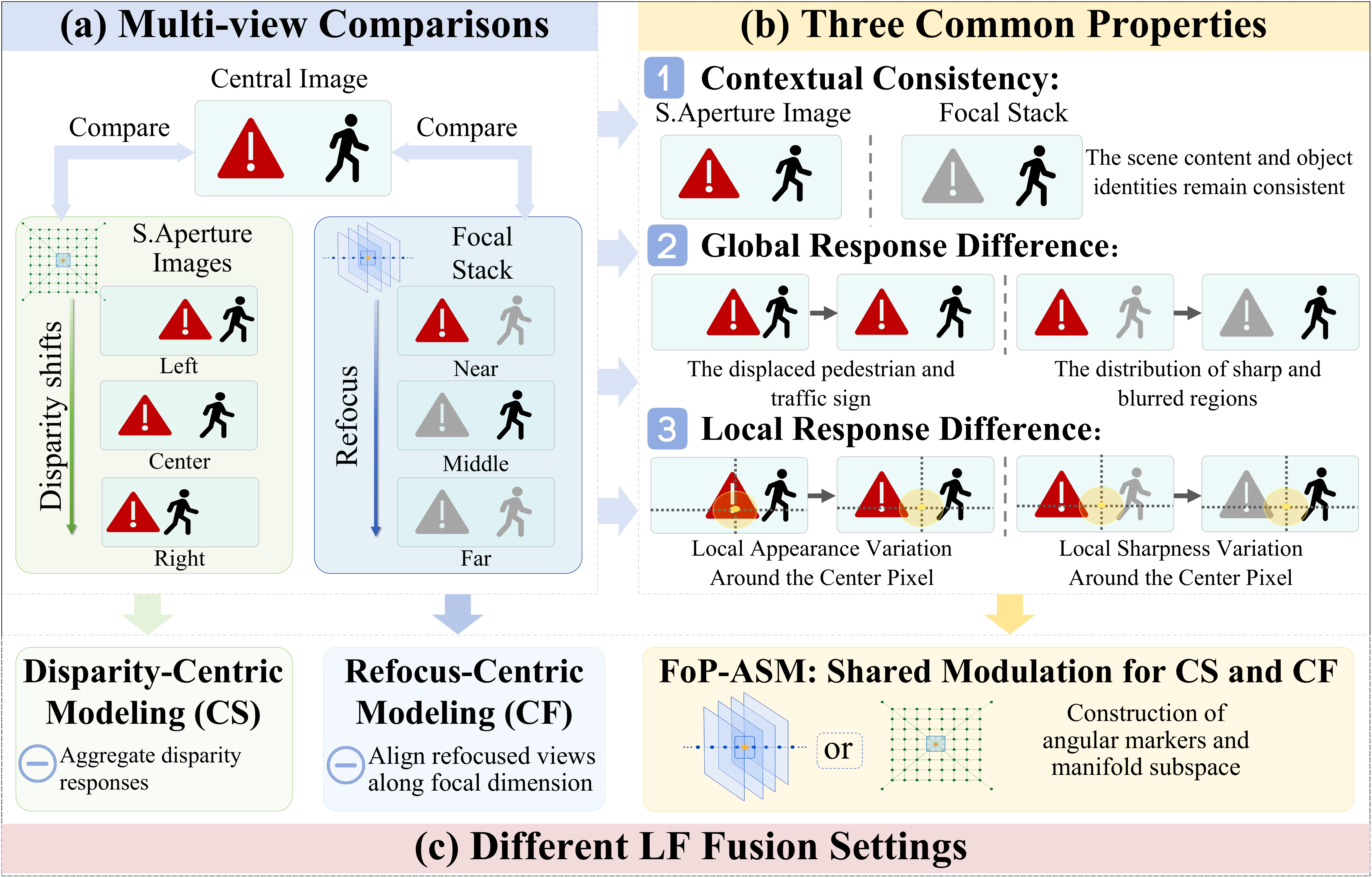}
    \caption{\changedv{Comparison of LF observation forms and modeling settings.
(a) S.Aperture images introduce viewpoint-induced disparity shifts, whereas FS
contains refocused observations at different focal planes. (b) Across different LF forms, comparing each auxiliary observation with the central image reveals three shared properties: contextual consistency, global response difference, and local response difference. (c) Existing form-specific methods model auxiliary-to-auxiliary
disparity or focus relationships, whereas LFX can be instantiated and trained separately under either CS or
CF.}}
    \label{fig:compare}
\vspace{-1em}
\end{figure}

\changedv{Despite these differences, comparing each auxiliary observation
independently with the central image reveals three properties shared
by both LF forms, as illustrated in Fig.~\ref{fig:Fig12}(b). First, contextual consistency indicates that the scene content and object identities remain consistent despite viewpoint changes in S.Aperture images or focal-plane changes in FS. 
Second, global response difference captures changes distributed across broad image regions. In S.Aperture images, disparity shifts change the overall spatial arrangement of the pedestrian and traffic sign. In FS, focal-plane changes alter the distribution of sharp and blurred responses over the corresponding objects. 
Third, local response difference captures changes concentrated at specific spatial positions. Within the yellow-highlighted regions, disparity shifts change the scene content around the center position in S.Aperture images, whereas focal-plane changes alter the response of the center position relative to its surroundings in FS.} \changedv{Fig.~\ref{fig:compare} compares modeling designs for the two
LF forms.
In Fig.~\ref{fig:compare}(a), each auxiliary observation is compared
independently with the central image.
The three shared properties in Fig.~\ref{fig:compare}(b), panels~1--3,
provide a common basis for modeling both LF forms.
Fig.~\ref{fig:compare}(c) contrasts existing form-specific designs
with the form-flexible design of LFX.
Despite progress in LF perception, a modeling design that can be
instantiated and trained separately under either CS or CF remains
underexplored.}

\changed{To address this gap, we establish LFX, a form-flexible LF perception framework that can be independently applied to either CS or CF under the same modeling paradigm. 
\changed{LFX decouples individual auxiliary observations from one another and applies a shared feature-modulation mechanism to them}.} Specifically, we propose \textbf{Field-of-Parallax Angular Subspace Modeling (FoP-ASM)}. \changed{FoP-ASM maps the feature difference between each auxiliary observation and the central image to an observation-specific angular marker. \changedv{The angular marker jointly encodes the global response difference,
which summarizes observation-level variation, and the local response
difference, which captures spatially localized variation.} This observation-wise construction avoids pairwise modeling among auxiliary observations and form-specific geometric assumptions.}
The marker is then injected into the corresponding auxiliary-observation feature to encode observation-specific angular characteristics. We further introduce manifold subspace modeling, which constrains angular residuals through manifold constraints and regularization losses, guiding cross-observation variations into a shared feature space while preserving \changedv{contextual consistency} across auxiliary observations.

Extensive evaluations on three LF perception datasets covering two LF forms demonstrate the effectiveness of LFX. 
On DUTLF-V2~\cite{DUTLFv2}, LFX achieves the best MAE values of $0.029$ and $0.027$ under the S.Aperture images and FS settings, reducing MAE by $12.12\%$ and $20.59\%$, respectively. On HFUT~\cite{zhang2020light}, LFX achieves MAE values of $0.061$ and $0.048$, corresponding to improvements of $6.15\%$ and $7.69\%$. We further validate LFX on the semantic segmentation task. 
On UrbanLF-real~\cite{urbanlf}, LFX achieves the best mIoU of $83.77\%$ and $79.89\%$ under the two settings, outperforming the previous best form-specific methods by $0.66\%$ and $2.52\%$, respectively.

\noindent The major contributions of this work are:
\begin{compactitem}
\item A Form-Flexible LF Modeling Framework (LFX). We propose LFX, a form-flexible LF perception framework. The same modeling design can be instantiated and trained separately under two fusion settings: the central image with S.Aperture images and the central image with FS.

\item Field-of-Parallax Angular Subspace Modeling (FoP-ASM). 
\changed{We propose FoP-ASM, which \changedv{encodes the global response difference and local response
difference of each auxiliary observation relative to the central image} as an angular marker and constrains the resulting residuals within a shared manifold subspace, without pairwise modeling among auxiliary observations.}

\item Cross-Form and Cross-Task Evaluation. We evaluate LFX on heterogeneous LF forms for salient object detection and further extend it to the semantic segmentation task. LFX achieves state-of-the-art performance on three LF datasets under two distinct observation forms.
\end{compactitem}

\section{Related Work}
The utilization of 4D LF cameras for scene understanding represents a rapidly evolving domain that holds significant potential for spatial intelligence~\cite{ar1,spatialintelligence,spatialintelligence2}. 
\changed{However, existing research typically develops modeling designs for a single LF observation form. Therefore, a form-flexible modeling design that can be instantiated and trained separately under either CS or CF remains underexplored.}
Therefore, we focus on three main aspects: disparity-centric and refocus-centric forms, and feature adaptation strategies.

\vspace{-0.5em}
\subsection{Disparity-centric form} 
In LF imaging, disparity variance provides multi-view observations~\cite{Disparity1,Disparity2,Disparity3,zhang2020multi}. Through multi-view spatial feature alignment, networks can learn cross-view consistent structural representations and boundary-aware geometric relationships.
PANet~\cite{parallex1} interprets different S.Aperture images as moving viewpoints and converts them into an optical-flow modality for salient object detection.
\changedv{CDINet~\cite{sparse} models correlations among multiple S.Aperture images through complementary and discriminative interactions and employs low-rank decomposition to identify informative LF views for sparse-view salient object detection.} 
LFIENet~\cite{combine} and LFIENet++~\cite{cong2024end} adopt a two-stage strategy that first estimates disparity maps from S.Aperture images and then projects them into depth-aware representations for semantic segmentation. 
While LFIENet-based methods are effective for disparity-centric LF inputs, they require costly pre-processing to generate disparity maps. In summary, these methods rely on viewpoint-dependent geometric assumptions and fail to accommodate refocus-centric form.

\vspace{-0.3em}
\subsection{Refocus-centric form} 
LF refocusing~\cite{lightrefocus} generates a sequence of auxiliary observations focused at different focal levels, providing effective focus information that has been widely exploited in prior work.
TENet~\cite{tenet} and LFTracy~\cite{lftracy} employ tailored mechanisms to fuse features from FS at multiple focal levels, implicitly integrating focus information with contextual cues.
\changedv{LFTransNet~\cite{liu2023lftransnet} introduces a learnable weight descriptor to adaptively aggregate focus information across FS observations.} 
FESNet~\cite{fesnet} introduces a joint learning framework with skip connections that transfer encoder information to the decoder, mitigating the loss of focal information during feature propagation. 
LFSODNet~\cite{zheng2024spatial} further treats FS as depth cues and emphasizes extracting spatial information from the central image. 
Nevertheless, these approaches depend on focal information obtained under approximately planar-aligned conditions and lack explicit modeling of the relative displacement and disparity structural variations across observations.

\changed{Overall, existing methods have advanced LF perception by modeling
viewpoint-disparity and focus-response relationships, respectively. However,
they are typically tailored to the pairwise relationships within a single
observation form.} A modeling design that can be instantiated and trained
separately under either CS or CF therefore remains underexplored. To address
this gap, we develop LFX.

\subsection{Feature Adaptation Strategies} 
Feature adaptation and parameter-efficient tuning have been extensively explored to transfer \changed{pre-trained visual encoders~\cite{he,Segformer,backbone3,backbone4}}, including CNN- and Transformer-based backbones, to downstream perception tasks~\cite{Fine_tuning1,VP1,VP2}. Existing adaptation methods typically insert lightweight modules into pre-trained backbones to refine intermediate representations, enabling multimodal feature reconfiguration and complementary information mining. Specifically, methods~\cite{VP3,VP5,VP6} introduce additional adapter modules into vision transformers (ViTs) to inject image-related inductive biases and recover fine-grained multi-scale representations, making ViT encoders more suitable for dense prediction tasks. Methods~\cite{cmx,zhang2020revisiting,cmnext,wang2023sgfnet} introduce a feature rectification mechanism into multimodal scene understanding, where one modality is leveraged to calibrate the other and facilitate long-range contextual interaction, enabling effective feature adaptation across different modalities. 
ST-Adapter~\cite{VP4} incorporates compact spatio-temporal adaptation modules into image-pretrained models, equipping the encoder with temporal reasoning ability for video action recognition. 
Despite their success in natural image understanding and multimodal learning, existing adaptation strategies are still primarily tailored to complementary or modality-aligned inputs. Unlike conventional 2D$/$3D images, LF data inherently encode four-dimensional spatial-angular information~\cite{lfimaging,lfimaging2}. Therefore, effectively exploiting the information embedded in different LF forms during feature encoding remains an important yet relatively underexplored direction in LF perception.

\section{Method}
This section presents LFX, a \changed{form-flexible} framework for light field perception. We formulate two representative LF forms and examine their representational discrepancies and shared properties from disparity-centric and refocus-centric perspectives in Sec.~\ref{sec:3.1} and Sec.~\ref{sec:RIS}. 
Meanwhile, the overall architecture of LFX is introduced in Sec.~\ref{sec:3.2}. 
Furthermore, we detail the Field-of-Parallax Angular Subspace Modeling (FoP-ASM) module in Sec.~\ref{sec:3.3}.

\subsection{Problem Formulation}\label{sec:3.1}
The angular resolution $(u,v)$ combined with the planar representation $(x,y)$ offers diverse image forms that enhance the LF perception model performance.

\noindent\textbf{Disparity-centric form.}
The disparity-centric form preserves pixel-wise variations across viewpoints by explicitly maintaining the angular sampling structure of the LF. 
Given angular resolution $(u,v)$ on the aperture plane, the discrete set of auxiliary observations (S.Aperture images) can be viewed as samples of the continuous 4D radiance field, forming an angular stack with cardinality $n = |u| \times |v|$.
Encoding is performed over the spatial domain $(x,y)$, thereby compressing the four-dimensional LF into a geometry-aware three-dimensional feature representation $S(L(n,x,y))$ that preserves disparity-induced structural variations.

\noindent\textbf{Refocus-centric Form.}
LF refocusing is fundamentally an angular integration process associated with a virtual focal plane. Given a 4D LF $L(u,v,x,y)$, synthetically shifting the imaging system to a virtual focus distance $d$ requires applying a shear transformation in epipolar geometry:
\begin{align}
(x', y') = \big(x + \alpha(d)\,u,\; y + \alpha(d)\,v\big).
\end{align}
Here, $\alpha(d)$ denotes a shear coefficient determined by the position of the virtual focal plane, which characterizes the equivalent spatial displacement induced by angular coordinates.
Specifically, the term $\alpha(d)u$ (and similarly $\alpha(d)v$) represents the proportional lateral translation on the image plane required for rays originating from different aperture locations to become spatially aligned when refocused onto the target focal plane. After shear-based alignment, integrating over the angular domain yields a refocused image at plane $d$:
\begin{align}
I_d(x,y)=\iint L(u,v,x',y')\,du\,dv.
\end{align}
This process allows the focal plane of the imaging system to be shifted to an arbitrary virtual focus distance without altering the physical lens configuration.
Repeating this operation over different virtual focus distances produces a set of refocused auxiliary observations (\textit{i.e.}, focal stacks), which provide explicit focus-dependent cues for downstream perception tasks.

\subsection{Shared Structure Across LF Observation Forms}
\label{sec:RIS}
To construct a form-flexible perception framework, we analyze how S.Aperture images and FS relate to their corresponding central image. \changedv{As illustrated in Fig.~\ref{fig:compare}(a)--(b), both LF forms exhibit three shared properties. \textit{Contextual consistency:} Although S.Aperture images and FS are generated through different LF formation processes, they preserve the same scene content and object semantics as the central image. As shown in the first row of Fig.~\ref{fig:compare}(b), the traffic sign and pedestrian remain present despite viewpoint-induced displacement in S.Aperture images and focal-plane-dependent focus changes in FS. \textit{Global response difference:} Global response difference describes changes distributed across the whole image between each auxiliary observation and the central image. As shown in the second row of Fig.~\ref{fig:compare}(b), viewpoint changes displace the pedestrian and traffic sign in S.Aperture images, whereas focal-plane changes alter the distribution of sharp and blurred object regions in FS. \textit{Local response difference:} Local response difference describes changes within specific spatial neighborhoods between each auxiliary observation and the central image. As shown in the third row of Fig.~\ref{fig:compare}(b), disparity shifts change the scene content surrounding the highlighted center position in S.Aperture images, whereas focal-plane changes alter the sharpness of the object content surrounding the corresponding position in FS. These shared properties reveal a common central-reference structure across the two LF forms. Building on this structure, LFX adopts an observation-wise design that can be instantiated and trained separately under either CS or CF. Specifically, FoP-ASM constructs an observation-specific angular marker from the global and local response differences of each auxiliary observation, while manifold subspace modeling promotes contextual consistency.}

\subsection{LFX Framework} \label{sec:3.2}

\begin{figure*}[t]
    \centering
    \includegraphics[width=0.90\linewidth]{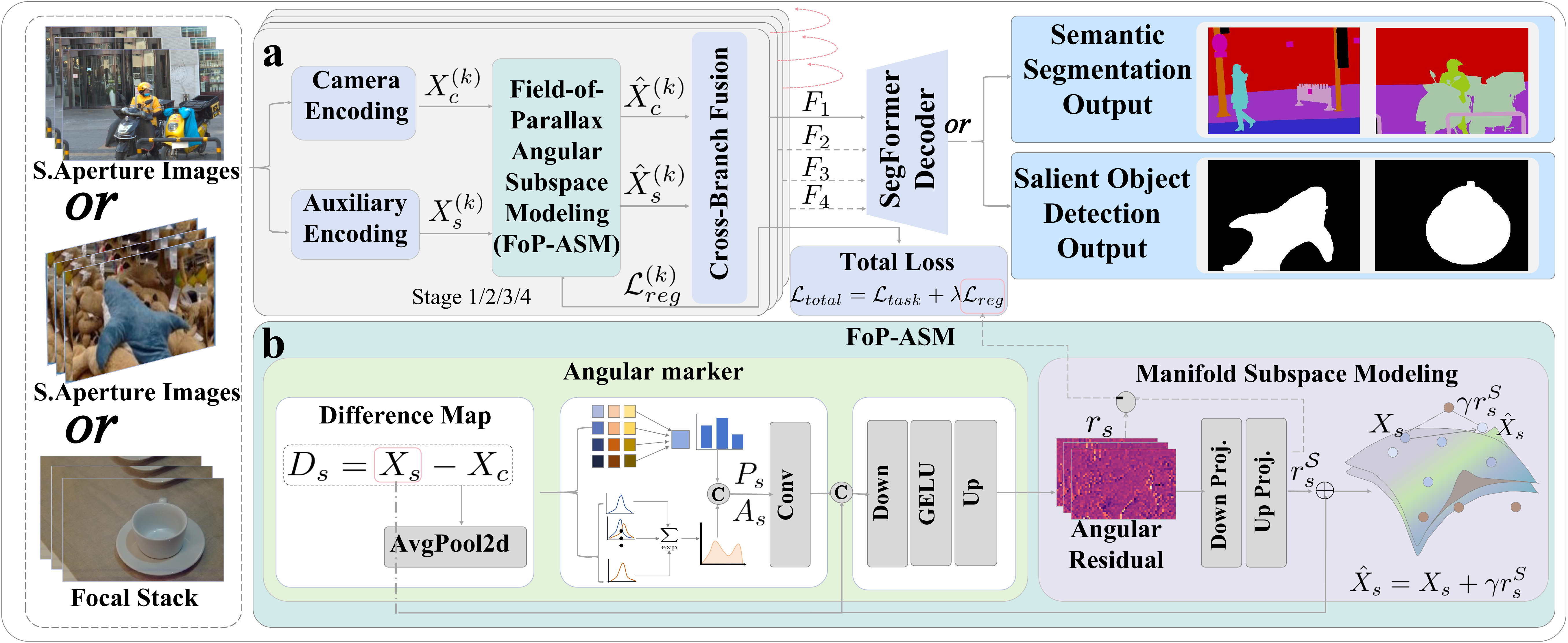}
    \caption{Overall framework of LFX. (a) illustrates the encoding stages of LFX, while (b) presents the architecture of the proposed FoP-ASM module. By independently modeling each auxiliary image and enforcing contextual consistency, our framework supports heterogeneous LF forms and achieves state-of-the-art performance on both salient object detection and semantic segmentation tasks.}
    \label{fig:framework}
    \vspace{-1em}
\end{figure*}

As illustrated in Fig.~\ref{fig:framework}, FoP-ASM serves as a modulator between backbone encoding and cross-branch fusion. 
Let $k$ denote the $k$-th backbone stage and $C_k$ denote the channel dimension at that stage. 
At each stage, the camera branch first undergoes the encoding block, producing the stage feature as $X_c^{(k)} \in \mathbb{R}^{C_k \times H_k \times W_k}$. The auxiliary branch features are processed by the corresponding downsampling layer, yielding:
\begin{equation}
\{X_s^{(k)}\}_{s=2}^{S}, 
\quad 
X_s^{(k)} \in \mathbb{R}^{C_k \times H_k \times W_k}.
\end{equation}
Thus, at stage $k$, we obtain a total of features 
$\{X_s^{(k)}\}_{s=1}^{S},$
where $X_1^{(k)} = X_c^{(k)}$ denotes the central image feature and the remaining $\{X_s^{(k)}\}_{s=2}^{S}$ correspond to auxiliary image features. 
These features are jointly fed into FoP-ASM. 

\changed{The module constructs angular markers through observation-wise differences and performs FoP-guided subspace modulation, producing $\{\hat{X}_s^{(k)}\}_{s=1}^{S}$, and the stage-wise regularization term $\mathcal{L}_{reg}^{(k)}$.} \changed{The features $\{\hat{X}_s^{(k)}\}_{s=1}^{S}$ are subsequently processed by the Feature Rectification Module (FRM) and Feature Fusion Module (FFM) inherited from CMNeXt~\cite{cmnext} and CMX~\cite{cmx}, producing the stage-wise fused feature $F_k$. The fused features from the four encoder stages, $\mathcal{F}=\{F_1,F_2,F_3,F_4\}$, are then passed to the SegFormer decoder for multi-scale feature aggregation~\cite{cmnext,Segformer}. Finally, the task-specific prediction layer maps the decoded feature to $14$-channel class logits for semantic segmentation or a single-channel saliency logit map for salient object detection.}

\subsection{Field-of-Parallax Angular Subspace Modeling (FoP-ASM)}\label{sec:3.3}
To accommodate heterogeneous LF forms, we decouple auxiliary images and assign each image an individual angular marker, enabling the model to capture observation-specific variations while preserving semantic consistency. \changed{We then describe the construction of each FoP-ASM component in detail. The implementation procedure is shown in
Algorithm~\ref{alg:r3c2-fopasm}.}

\noindent\textit{(1) \changed{Construction of the Angular Marker}:} Given multi-observation features: 
\begin{equation}
\{X_s\}_{s=1}^{S}, \quad 
X_s \in \mathbb{R}^{C \times H \times W},
\end{equation}
where $S$ denotes the number of observations, $C$ the channel dimension, and $H \times W$ the spatial resolution. Taking the central image feature $X_1=X_c$ as the reference, we compute the feature difference between each auxiliary observation and the central image:
\begin{equation}
D_s = X_s - X_c, \quad s=2,\ldots,S,
\end{equation}
where $X_s$ denotes the feature of the $s$-th auxiliary image. 
\changedv{We further decompose $D_s$ into two complementary response maps.
The global response difference $P_s$ summarizes the overall feature variation
relative to the central image, whereas the local response difference $A_s$
emphasizes strong channel-wise variations at individual spatial positions.} 
After applying 2D average pooling to $D_s$ as a spatial low-pass filter, we
obtain \changedv{the global response difference $P_s$} by averaging the filtered difference map
along the channel dimension. \changed{We then aggregate the difference map along the channel dimension to extract \changedv{the local response difference $A_s$}.}  
\begin{equation}
P_s = \frac{1}{C} \sum_{c=1}^{C} D_s^{(c)},
\quad
A_s = \tau \log \sum_{c=1}^{C} 
\exp\left(\frac{D_s^{(c)}}{\tau}\right).
\end{equation} \label{eq1}
The $\tau$ is a temperature parameter (0.07 following \cite{chen2020simple,wang2021understanding}).  $P_s$ and $A_s$ are computed for each auxiliary observation ($s \in \{2,\ldots,S\}$). The {angular embedding is obtained as:}
\begin{equation}
\Phi_s = \operatorname{Concat}(P_s,A_s),
\quad
\Phi_s \in \mathbb{R}^{2 \times H \times W}.
\end{equation}
\changed{This embedding serves as an \textit{angular marker} for each observation, characterizing its relative position and directional variation. The original feature $X_s$ ($s \in \{2,\ldots,S\}$) is then conditioned on its angular embedding $\Phi_s$.} Then, we project the embedding:
\begin{equation}
Z_s = W_q(\Phi_s),
\quad
Z_s \in \mathbb{R}^{d \times H \times W},
\end{equation}
and generate a conditional angular residual $r_s$ ($s=2,\ldots,S$):
\begin{equation}
r_s = W_{up}\!\left(\sigma\!\left(W_{down}([Z_s, X_s])\right)\right),
\quad
r_s \in \mathbb{R}^{C \times H \times W},
\end{equation}
where $W_{down}$ and $W_{up}$ denote the down- and up-projection layers for angular marker fusion, respectively.
This encodes FoP-guided angular modulation in feature space. 

\noindent\textit{(2) Manifold subspace modeling:} To enforce semantic consistency across heterogeneous LF forms, we introduce a shared manifold subspace. 
Specifically, given the angular residual $r_s\in\mathbb{R}^{C\times H\times W}$ for $s\in\{2,\ldots,S\}$, and central image feature $X_c$, \changed{we first form the input collection \changed{$\{r_s^*\}_{s=1}^{S}=\{X_c,r_2,\ldots,r_S\}$}, project it into a compact latent subspace, and then reconstruct it back to the original channel dimension:}
\begin{equation}
r_s^{\mathcal{S}} =
W_{up}^{\mathcal{S}}
\!\left(
W_{down}^{\mathcal{S}}(r_s^*)
\right),
\quad
r_s^{\mathcal{S}} \in \mathbb{R}^{C \times H \times W}, 
\end{equation}
\changed{where $r_1^*=X_c$ denotes the central image feature, and $r_s^*=r_s\in\mathbb{R}^{C\times H\times W}$ denotes the angular residual of the $s$-th auxiliary observation for $s\in\{2,\ldots,S\}$.}  $W_{down}^{\mathcal{S}}$ and $W_{up}^{\mathcal{S}}$ denote the down- and up-projection operators of the shared manifold subspace, respectively. It is worth noting that, unlike the previous stage that introduces angular cues only to the auxiliary observations ($s \in \{2,\ldots,S\}$), this stage jointly feeds the angular residuals and the central image feature into a shared projection space. 
The purpose is to establish implicit feature-level dependencies between the central image and auxiliary observations, and to jointly constrain their consistency within a shared semantic space. 
After that, a dropout operation is applied to improve regularization:
\begin{equation}
\bar{r}_s^{\mathcal{S}}=\mathrm{Dropout}\left(r_s^{\mathcal{S}}\right),
\end{equation}
where $s \in \{1,\ldots,S\}$. When the residual low-pass branch is enabled, $\bar{r}_s^{\mathcal{S}}$ is  processed by an average-pooling low-pass filter:
\begin{equation}
\tilde{r}_s^{\mathcal{S}}
=
\mathrm{AvgPool}_{k_r\times k_r}
\left(
\bar{r}_s^{\mathcal{S}}
\right),
\end{equation}
where $k_r$ denotes the kernel size of the residual low-pass filter. 
The final feature update is performed in a residual manner:
\begin{equation}
\hat{X}_s = X_s + \gamma \tilde{r}_s^{\mathcal{S}},
\end{equation}
where $\gamma$ is a learnable scaling factor. 
Finally, $\hat{X}_1$ is used as the updated main feature, and the remaining $S-1$ representations are returned as auxiliary LF features. This preserves the original semantic content while maintaining FoP-consistent angular modulation.

\noindent\textit{(3) Subspace Regularization:} To further promote contextual consistency, we impose a regularization term that penalizes deviations from the learned subspace:
\begin{equation}
\mathcal{L}_{reg}
=
\frac{1}{|\Omega|}
\sum_{s=1}^{S}
\left\|
r_s^*-r_s^{\mathcal{S}}
\right\|_F^2,
\end{equation}
where $|\Omega| = S C H W$ \changed{denotes the total number of elements in the shared-subspace input collection}. The loss is:
\begin{equation}
\mathcal{L}_{total}
=
\mathcal{L}_{task}
+
\lambda \mathcal{L}_{reg},
\end{equation}
where $\lambda$ balances task supervision and context alignment. For the four-stage backbone in our framework, the regularization term $\mathcal{L}_{reg}$ is computed as the summation of the stage-wise regularization losses.

\begin{algorithm}[!t]
\SetAlCapNameFnt{\footnotesize}
\SetAlCapFnt{\footnotesize}
\SetAlCapSkip{3pt}
\caption{FoP-ASM Procedure}
\label{alg:r3c2-fopasm}

\scriptsize
\linespread{1.30}\selectfont
\SetAlgoLined
\DontPrintSemicolon

\KwIn{\parbox[t]{0.78\linewidth}{\raggedright\changed{%
$\{X_s\}_{s=1}^{S}$, where
$X_s\in\mathbb{R}^{C\times H\times W}$;
$X_1=X_c$ is the central image feature, and
$\{X_s\}_{s=2}^{S}$ are the auxiliary image features.}}}

\KwOut{\parbox[t]{0.78\linewidth}{\raggedright\changed{%
Updated central image feature $\hat X_1$, updated auxiliary features
$\{\hat X_s\}_{s=2}^{S}$, and regularization loss
$\mathcal L_{reg}$.}}}

\textbf{(1) Construction of the Angular Marker}\;

$X_c\leftarrow X_1$\;

\For{$s\gets2$ \KwTo $S$}{
    $D_s\leftarrow X_s-X_c$\;

    $P_s\leftarrow C^{-1}\sum_{c=1}^{C}D_s^{(c)}$,\quad
    $A_s\leftarrow\tau\log\sum_{c=1}^{C}
    \exp\!\left(D_s^{(c)}/\tau\right)$\;

    $\Phi_s\leftarrow[P_s,A_s]_{\mathrm{cat}}$,\quad
    $Z_s\leftarrow W_q(\Phi_s)$\;

    $r_s\leftarrow
    W_{up}\!\left(
    \sigma\!\left(
    W_{down}([Z_s,X_s]_{\mathrm{cat}})
    \right)
    \right)$\;
}

\textbf{(2) Manifold Subspace Modeling}\;

\changed{
$\{r_s^*\}_{s=1}^{S}
\leftarrow
\{X_c,r_2,\ldots,r_S\}$
}\;

\For{$s\gets1$ \KwTo $S$}{
    \changed{
    $r_s^{\mathcal S}\leftarrow
    W_{up}^{\mathcal S}\!\left(
    W_{down}^{\mathcal S}(r_s^*)
    \right)$
    }\;

    \changed{
    $\bar r_s^{\mathcal S}\leftarrow
    \operatorname{Dropout}(r_s^{\mathcal S})$
    },\quad
    \changed{
    $\tilde r_s^{\mathcal S}\leftarrow
    \operatorname{AvgPool}_{k_r\times k_r}
    (\bar r_s^{\mathcal S})$
    }\;

    $\hat X_s\leftarrow
    X_s+\gamma\tilde r_s^{\mathcal S}$\;
}

\textbf{(3) Subspace Regularization}\;

$\mathcal L_{reg}\leftarrow
|\Omega|^{-1}\sum_{s=1}^{S}
\left\|r_s^*-r_s^{\mathcal S}\right\|_F^2$\;

\KwRet{\changed{
$\hat X_1,\{\hat X_s\}_{s=2}^{S},\mathcal L_{reg}$
}}\;

\end{algorithm}
\vspace{-2em}

\section{Experiments}

\begin{table}[t]
    \centering
    \caption{Results on DUTLF-V2~\cite{DUTLFv2} under S.Aperture images and FS
    forms. ``$\uparrow$'' indicates that a higher value is preferred,
    ``$\downarrow$'' indicates that a lower value is preferred, and
    ``A$\mathrm{vg.R}$'' indicates the average ranking. 
    Avg.R and Gain are computed among LF methods
    within each form, excluding the 2D methods.}
    \label{tab:duft-single}
    \footnotesize
    \renewcommand{\arraystretch}{1.10}
    \setlength{\tabcolsep}{1.2pt}
    \begin{tabular*}{\columnwidth}{@{\extracolsep{\fill}}lccccc@{}}
        \toprule
        Method & MAE $\downarrow$ & E $\uparrow$ & S $\uparrow$
               & F $\uparrow$ & Avg.R $\downarrow$ \\
        \midrule
        \multicolumn{6}{c}{\textit{2D Methods}} \\
        \midrule
        BBRF~\cite{boosting} & 0.051 & 0.907 & 0.869 & 0.843 & -- \\
        LeNo~\cite{leno} & 0.047 & 0.922 & 0.869 & 0.868 & -- \\
        ConTriNet~\cite{divide} & 0.057 & 0.903 & 0.847 & 0.806 & -- \\
        MDSAM~\cite{multi} & 0.044 & 0.940 & 0.940 & 0.898 & -- \\
        \midrule
        \multicolumn{6}{c}{\textit{S.Aperture images}} \\
        \midrule
        ESCNet~\cite{exploring} & 0.042 & 0.929 & 0.881 & 0.850 & 5.88 \\
        OBGNet~\cite{occlusion} & 0.037 & 0.936
            & \textcolor{pink}{0.896} & 0.869 & 3.38 \\
        DLGLRG~\cite{lightt} & 0.046 & 0.908 & 0.861 & 0.816 & 8.50 \\
        \changed{MTCNet}~\cite{zhang2020multi} & 0.039 & 0.932
            & 0.895 & 0.866 & 4.50 \\
        LFDCN~\cite{zhang2020light} & 0.058 & 0.900 & 0.867 & 0.799 & 9.75 \\
        MoLF~\cite{memory} & 0.047 & 0.912 & 0.876 & 0.811 & 8.25 \\
        DLFS~\cite{deep} & 0.076 & 0.875 & 0.809 & 0.752 & 11.00 \\
        CDINet~\cite{sparse} & \textcolor{pink}{0.033}
            & \textcolor{pink}{0.939} & \textcolor{violet}{0.905}
            & 0.880 & \textcolor{pink}{2.25} \\
        \changed{UGRAN}~\cite{ugran} & \changed{0.054} & \changed{0.928}
            & \changed{0.877} & \changed{0.893} & 6.25 \\
        \changed{USCNet}~\cite{uscnet} & \changed{0.042} & \changed{0.917}
            & \changed{\textcolor{pink}{0.896}}
            & \changed{\textcolor{pink}{0.901}} & 4.25 \\
        LFX (Ours) & \textcolor{violet}{0.029}
            & \textcolor{violet}{0.941} & 0.890
            & \textcolor{violet}{0.922} & \textcolor{violet}{2.00} \\
        \midrule
        Gain & $+12.12\%$ & \changed{$+0.21\%$} & \changed{$-1.66\%$}
             & \changed{$+2.33\%$} & -- \\
        \midrule
        \multicolumn{6}{c}{\textit{FS}} \\
        \midrule
        MoLF$_{12}$~\cite{memory} & 0.064 & 0.866 & 0.825 & 0.724 & 8.38 \\
        SANet$_{12}$~\cite{zhang2021learning} & 0.065 & 0.903 & 0.866 & 0.834 & 6.75 \\
        MEANet$_{12}$~\cite{meanet} & 0.056 & 0.895 & 0.843 & 0.784 & 7.25 \\
        GFRNet$_{12}$~\cite{guided} & \textcolor{pink}{0.034}
            & 0.922 & 0.891 & 0.861 & 3.63 \\
        TLFNet$_{12}$~\cite{transformer} & 0.038
            & \textcolor{pink}{0.933} & 0.890 & 0.859 & 3.50 \\
        Sigma$_{12}$~\cite{wan2025sigma} & 0.077 & 0.929 & 0.825 & 0.815 & 7.25 \\
        \changed{UGRAN}~\cite{ugran} & \changed{0.048} & \changed{0.929}
            & \changed{\textcolor{pink}{0.897}} & \changed{0.853} & 4.13 \\
        \changed{USCNet}~\cite{uscnet} & \changed{0.040}
            & \changed{\textcolor{violet}{0.934}} & \changed{0.891}
            & \changed{\textcolor{pink}{0.928}} & \textcolor{pink}{2.63} \\
        LFX (Ours) & \textcolor{violet}{0.027} & 0.931
            & \textcolor{violet}{0.910} & \textcolor{violet}{0.940}
            & \textcolor{violet}{1.50} \\
        \midrule
        Gain & $+20.59\%$ & \changed{$-0.32\%$} & \changed{$+1.45\%$}
             & \changed{$+1.29\%$} & -- \\
        \bottomrule
    \end{tabular*}
    \vspace{-1em}
\end{table}

\subsection{Datasets and Implementation Details}

\noindent\textbf{Datasets.} We conduct salient object detection on DUTLF-V2~\cite{DUTLFv2} and HFUT~\cite{zhang2020light} under two LF forms, \changed{namely focal stack (FS) and sub-aperture images (S.Aperture images)}. 
To further verify the effectiveness of the proposed method, we extend it to the semantic segmentation task and evaluate it on the UrbanLF-real dataset~\cite{urbanlf}. 
All three datasets adopt a $9 \times 9$ angular resolution. 
For the FS form, each sample contains up to $12$ refocused slices at different virtual focal depths. \noindent\textbf{Implementation Details.} 
All experiments are conducted on four NVIDIA RTX 3090 GPUs with a batch size of $1$ per GPU. 
\changed{For S.Aperture images, a $2 \times 2$ angular subset is sampled from the full $9 \times 9$ image, while $4$ refocused slices are used in the FS setting}. 
\changed{Unless otherwise specified, the models under the S.Aperture images and FS settings are trained and evaluated separately.} 
SegFormer-B4~\cite{Segformer} is used for the main experiments, while SegFormer-B2 is adopted for the ablation studies. 
For salient object detection, we report four evaluation metrics, including MAE, E-measure, S-measure, and F-measure, where MAE serves as the primary evaluation metric. For semantic segmentation, we report Acc, mAcc, and mIoU, with mIoU used as the principal metric.

\begin{table}[t]
    \centering
    \caption{Quantitative comparison on the HFUT~\cite{zhang2020light} dataset under S.Aperture
    images and FS forms. ``$\uparrow$'' indicates that a higher
    value is better, whereas ``$\downarrow$'' indicates that a lower value is
    better, and ``A$\mathrm{vg.R}$'' indicates the average ranking.}
    \label{tab:r1c1-hfut-comparison}
    \footnotesize
    \renewcommand{\arraystretch}{1.10}
    \setlength{\tabcolsep}{3.5pt}
    \begin{tabular*}{\columnwidth}
    {@{\extracolsep{\fill}}lccccc}
        \toprule
        Method & MAE $\downarrow$ & E $\uparrow$ & S $\uparrow$
        & F $\uparrow$ & Avg.R $\downarrow$ \\
        \midrule
        \multicolumn{6}{c}{\textit{S.Aperture images}} \\
        \midrule
        DLFS~\cite{deep}
        & 0.109 & 0.755 & 0.712 & 0.592 & 8.88 \\
        LFDCN~\cite{zhang2020light}
        & 0.109 & 0.797 & 0.730 & 0.671 & 7.75 \\
        OBGNet~\cite{occlusion}
        & 0.094 & 0.790 & 0.740 & 0.671 & 7.38 \\
        MTCNet~\cite{zhang2020multi}
        & 0.084 & 0.818 & 0.751 & 0.714 & 5.13 \\
        GAGNN~\cite{zhang2021geometry}
        & 0.069 & 0.844 & 0.798
        & \textcolor{pink}{0.762} & 3.25 \\
        ESCNet~\cite{exploring}
        & 0.086 & 0.814 & 0.751 & 0.682 & 5.88 \\
        PANet~\cite{parallex1}
        & \textcolor{pink}{0.065}
        & \textcolor{pink}{0.854}
        & 0.809 & 0.752
        & \textcolor{pink}{2.50} \\
        CDINet~\cite{sparse}
        & 0.072 & 0.847
        & \textcolor{pink}{0.823}
        & 0.749 & 3.25 \\
        \midrule
        LFX (Ours)
        & \textcolor{violet}{0.061}
        & \textcolor{violet}{0.951}
        & \textcolor{violet}{0.887}
        & \textcolor{violet}{0.896}
        & \textcolor{violet}{1.00} \\
        \midrule
        Gain
        & $+6.15\%$ & $+11.36\%$ & $+7.78\%$
        & $+17.59\%$ & -- \\
        \midrule
        \multicolumn{6}{c}{\textit{FS}} \\
        \midrule
        DLGLRG~\cite{lightt}
        & 0.071 & 0.841 & 0.766 & 0.709 & 8.50 \\
        SANet~\cite{zhang2021learning}
        & 0.080 & 0.821 & 0.781 & 0.723 & 9.25 \\
        PANet~\cite{piao2023panet}
        & 0.073 & 0.843 & 0.802 & 0.704 & 8.25 \\
        WSLF~\cite{liang2022weakly}
        & 0.097 & 0.794 & 0.727 & 0.652 & 11.00 \\
        FES-Trans~\cite{fesnet}
        & 0.062 & 0.872 & 0.846 & 0.800 & 4.25 \\
        FES-Net~\cite{fesnet}
        & 0.074 & 0.847 & 0.798 & 0.727 & 7.75 \\
        TENet~\cite{tenet}
        & 0.059 & 0.877 & 0.838 & 0.791 & 4.25 \\
        GFRNet~\cite{guided}
        & 0.072 & 0.852 & 0.803 & 0.756 & 6.25 \\
        STSA~\cite{PKU}
        & 0.057 & 0.879 & 0.834 & 0.805 & 3.50 \\
        FANet~\cite{fu2025fanet}
        & \textcolor{pink}{0.052}
        & \textcolor{pink}{0.891}
        & \textcolor{pink}{0.851}
        & \textcolor{pink}{0.843}
        & \textcolor{pink}{2.00} \\
        \midrule
        LFX (Ours)
        & \textcolor{violet}{0.048}
        & \textcolor{violet}{0.935}
        & \textcolor{violet}{0.885}
        & \textcolor{violet}{0.920}
        & \textcolor{violet}{1.00} \\
        \midrule
        Gain
        & $+7.69\%$ & $+4.94\%$ & $+4.00\%$
        & $+9.13\%$ & -- \\
        \bottomrule
    \end{tabular*}
    \vspace{-1em}
\end{table}

\subsection{\changed{Cross-form Evaluation on S.Aperture images and FS}} 
Tab.~\ref{tab:duft-single} presents the benchmark \changed{results on DUTLF-V2~\cite{DUTLFv2} under two LF forms}.  
\changed{Benefiting from the proposed form-flexible modeling framework}, LFX achieves the best performance on the primary evaluation metric (MAE) under both forms. Specifically, under the sub-aperture form, LFX reduces the MAE from $0.033$ to $0.029$, corresponding to a $12.12\%$ relative improvement over the previous best method. Under the FS form, the MAE is further reduced from $0.034$ to $0.027$, achieving a $20.59\%$ improvement compared with the previous method. 
In addition to the primary metric, LFX also improves several complementary evaluation metrics. \changed{Although LFX is slightly below the best result for S-measure under
S.Aperture images and E-measure under FS, it consistently achieves the best
average ranking across all compared LF methods, indicating strong overall
competitiveness.}
For reference, several RGB-D-based methods are also reported. 

\changed{Furthermore, we evaluate LFX on the HFUT dataset~\cite{zhang2020light}
under both S.Aperture images and FS, as reported in
Tab.~\ref{tab:r1c1-hfut-comparison}. LFX outperforms all compared methods
across MAE, E-measure, S-measure, and F-measure under both forms. Under
S.Aperture images, LFX reduces MAE by $6.15\%$ and improves E-measure,
S-measure, and F-measure by $11.36\%$, $7.78\%$, and $17.59\%$,
respectively, over the best competing results. Under FS, it reduces MAE by
$7.69\%$ and improves the other three metrics by $4.94\%$, $4.00\%$, and
$9.13\%$, respectively. LFX also achieves the best average ranking of $1.00$
under both forms.} Importantly, \changed{LFX achieves the best performance using the proposed form-flexible modeling framework}, whereas prior methods are tailored to a specific LF form.

\subsection{\changed{UrbanLF-real Dataset for S.Aperture images and FS}}
To evaluate the effectiveness of LFX,
\changed{we conduct the evaluation for semantic segmentation on UrbanLF-real under
both the CS and CF settings}. The quantitative results are summarized
in Tab.~\ref{table:urlablf-single}. Under the CS fusion setting, LFX achieves
the best mIoU of $83.77\%$, exceeding the previous best result of $83.22\%$
obtained by CMNeXt~\cite{cmnext}. \changed{The Acc of LFX is slightly lower than those of
OAFuser~\cite{oafuser}, CMNeXt~\cite{cmnext}, and
StitchFusion~\cite{stitchfusion}, whereas LFX achieves better mAcc and mIoU. This difference
among the metrics indicates that LFX does not rely solely on the large number
of pixels belonging to dominant classes to obtain high overall accuracy.} \changed{Meanwhile, following the five-view input configuration in~\cite{sparse}, LFX
uses one central image and four auxiliary S.Aperture images, yet still
outperforms CMNeXt~\cite{cmnext} and OAFuser~\cite{oafuser}, both of which use
nine input views. The effect of the number of auxiliary observations is
analyzed in Sec.~\ref{5_2}.} 
\changed{Compared with S.Aperture images, which provide observations from different
viewpoints, FS exhibits pixel-wise response variations caused by focal shifts,
posing a greater challenge to fine-grained feature discrimination.
Nevertheless, LFX achieves the best mIoU of $79.89\%$ under the CF fusion
setting.} Together with the performance improvements observed on the SOD task,
these results demonstrate the effectiveness of LFX across the two LF
observation forms.

\begin{table}[!t]
    \centering
    \caption{Semantic segmentation results on UrbanLF-real under S.Aperture images
    and FS forms. ``\textuparrow'' indicates that a higher
    value is preferred, and ``A$\mathrm{vg.R}$'' indicates the average
    ranking. 
    }
    \label{table:urlablf-single}
    \footnotesize
    \renewcommand{\arraystretch}{1.10}
    \setlength{\tabcolsep}{2pt}
    \begin{tabular*}{\columnwidth}{@{\extracolsep{\fill}}lcccc@{}}
        \toprule
        Model & Acc(\%) $\uparrow$ & mAcc(\%) $\uparrow$
              & mIoU(\%) $\uparrow$ & Avg.R \\
        \midrule
        \multicolumn{5}{c}{\textit{S.Aperture images}} \\
        \midrule
        Sigma~\cite{wan2025sigma} & 87.94 & 75.67 & 68.09 & 11.00 \\
        PSPNet-LF~\cite{zhao2017pyramid} & 91.75 & 84.31 & 77.79 & 10.00 \\
        OCR-LF~\cite{yuan2020object} & 92.92 & 86.50 & 80.69 & 6.67 \\
        LF-IENet$_5$~\cite{combine} & 92.40 & 84.94 & 79.03 & 9.00 \\
        LF-IENet$_4$~\cite{combine} & 93.02 & 86.69 & 80.35 & 6.33 \\
        LF-IENet$_5$++~\cite{cong2024end} & 92.70 & 85.74 & 79.69 & 8.00 \\
        LF-IENet$_4$++~\cite{cong2024end} & 93.34 & 87.13 & 81.09 & 5.00 \\
        OAFuser~\cite{oafuser} & \textcolor{violet}{94.45}
            & 88.21 & 82.69 & 3.00 \\
        CMNeXt~\cite{cmnext} & \textcolor{pink}{94.24}
            & 88.74 & \textcolor{pink}{83.22} & \textcolor{pink}{2.33} \\
        \changed{StitchFusion}~\cite{stitchfusion} & 94.02
            & \textcolor{pink}{88.82} & 83.11 & 2.67 \\
        \midrule
        LFX (Ours) & 93.69 & \textcolor{violet}{88.87}
            & \textcolor{violet}{83.77} & \textcolor{violet}{2.00} \\
        \midrule
    \multicolumn{5}{c}{\textit{FS}} \\
    \midrule
    \changed{SHIFNet}~\cite{zhao2025unveiling}
    & \changed{91.31} & \changed{84.51} & \changed{76.01} & \changed{4.33} \\
    \changed{SACNet}~\cite{wang2024alignment}
    & \changed{91.94}
    & \changed{\textcolor{pink}{85.58}}
    & \changed{\textcolor{pink}{77.93}}
    & \changed{\textcolor{pink}{2.33}} \\
    \changed{CMNeXt}~\cite{cmnext}
    & \changed{\textcolor{pink}{91.99}}
    & \changed{84.83} & \changed{77.16} & \changed{2.67} \\
    \changed{StitchFusion}~\cite{stitchfusion}
    & \changed{91.49} & \changed{83.37} & \changed{75.32} & \changed{4.67} \\
    \midrule
    \changed{LFX (Ours)}
    & \changed{\textcolor{violet}{92.65}}
    & \changed{\textcolor{violet}{86.46}}
    & \changed{\textcolor{violet}{79.89}}
    & \changed{\textcolor{violet}{1.00}} \\
        \bottomrule
    \end{tabular*}
    \vspace{-1em}
\end{table}

\begin{figure*}[ht]
  \centering
  \includegraphics[width=0.88\textwidth]{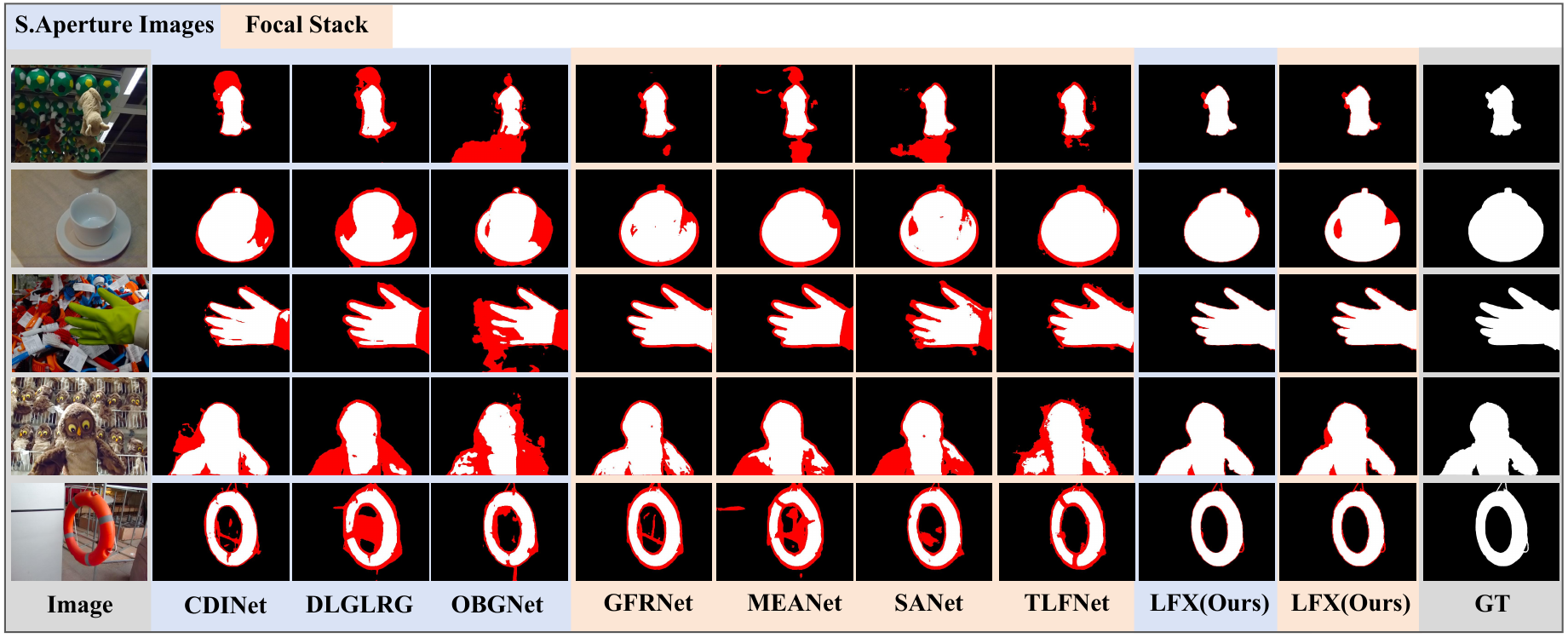}
\caption{
Qualitative visualization of difference maps for \changed{the SOD task in the DUTLF-V2~\cite{DUTLFv2} dataset.} In each difference map, \protect\textcolor[RGB]{255,0,0}{red regions} denote prediction errors, while white and black regions indicate correctly predicted regions. 
The \protect\textcolor[RGB]{80,80,80}{gray region} denotes the original image and ground truth, 
the \protect\textcolor[RGB]{40,80,160}{blue region} denotes visualization results of methods using S.Aperture images as input, 
and the \protect\textcolor[RGB]{180,80,40}{pink region} denotes visualization results of methods using FS images as input.}
\label{visualization1}
\vspace{-0.5em}
\end{figure*}

\begin{figure*}[ht]
  \centering
  \includegraphics[width=0.88\textwidth]{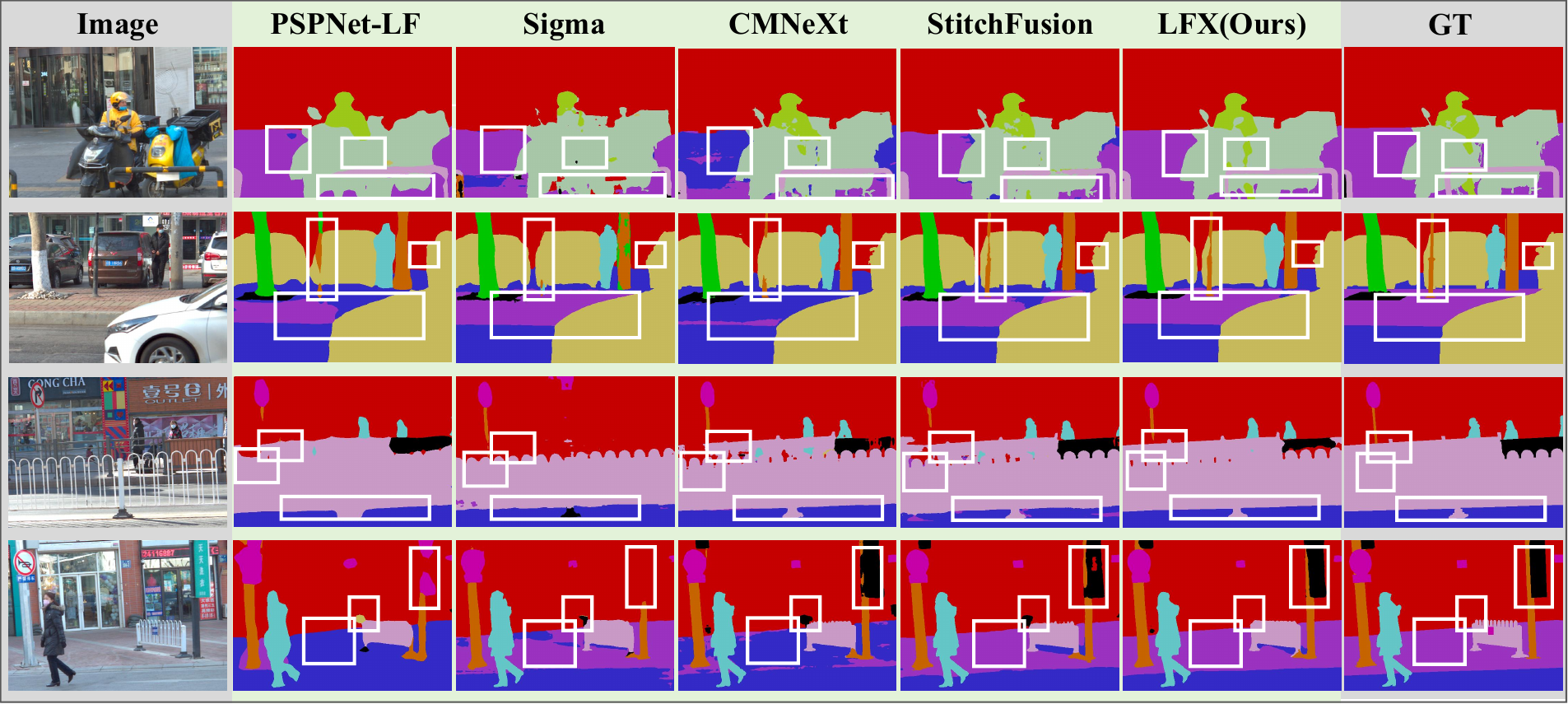}
    \caption{
    Qualitative comparison on the UrbanLF-real dataset~\cite{urbanlf} for the semantic segmentation task. Compared with existing methods, our method produces segmentation results with clearer boundaries and better semantic consistency.
    }
  \label{visualization2}
  \vspace{-1em}
\end{figure*}

\begin{figure*}[ht]
  \centering
  \includegraphics[width=0.88\textwidth]{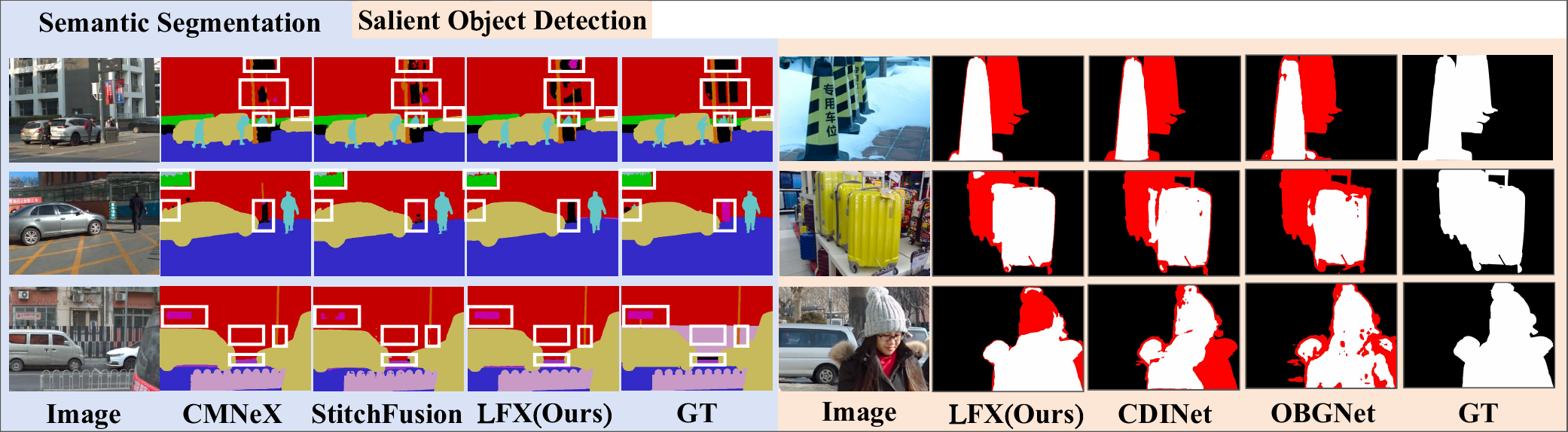}
  \caption{{Challenging examples for LF perception. 
  While our method demonstrates superior performance in capturing fine-grained structures compared to existing approaches, perception under complex lighting and environmental conditions remains highly challenging for all methods.}}
  \label{visualization3}
  \vspace{-1em}
\end{figure*}

\subsection{Qualitative visualization on different benchmarks.} 

\noindent\textbf{\changed{Results under S.Aperture images and FS forms for salient object detection.}} Fig.~\ref{visualization1} shows qualitative comparisons on \changed{the DUTLF-V2 dataset~\cite{DUTLFv2} under both S.Aperture images and FS forms.} We visualize the difference maps between predictions and ground truths for the same scenes across different methods and forms, where red regions indicate inconsistent predictions. Compared with those form-specific methods, LFX produces results with clearer object boundaries and a more consistent global context. \changed{Due to the inherent displacement bias in S.Aperture images, the results under this form are generally less stable.} 
However, by modeling each view independently, LFX effectively alleviates this issue and maintains consistent prediction quality across forms. \textbf{\changed{Results under S.Aperture images for semantic segmentation.}} To further validate the effectiveness of our approach, we extend LFX to the semantic segmentation task, leveraging S.Aperture images for fine-grained scene understanding. As shown in Fig.~\ref{visualization2}, our method accurately captures both global semantics and object boundaries (e.g., walls, trash bins, and guardrails). Compared with the baseline CMNeXt~\cite{cmnext}, LFX, with explicit modeling of angular features, achieves a more precise perception of objects in complex environments. \noindent\textbf{Challenging cases.} Furthermore, we present several challenging examples for semantic segmentation and salient object detection, as shown in Fig.~\ref{visualization3}. These cases highlight two key limitations of existing LF perception methods. First, environmental complexity: this includes complex structures, occlusions, and diverse object appearances. This remains challenging for robust LF understanding. Second, semantic limitation: most prior works adopt a closed-set paradigm, which restricts the network’s ability to accurately perceive rare object categories, such as hats. Our proposed LF framework effectively handles multiple LF forms, offering representational benefits for further research.

\section{Ablation Studies}
In this section, we present comprehensive ablation studies for LFX. The core component is evaluated in Sec.~\ref{5_1}, the number of input LF images in Sec.~\ref{5_2}, the internal designs of FoP-ASM in Sec.~\ref{5_3}, the angular-marker construction in Sec.~\ref{5_4}, and the efficiency comparison in Sec.~\ref{5_5}.

\begin{table*}[ht]
\centering
    \caption{Ablation study for the core components of LFX. ``M$\mathrm{ani.}$'' denotes the manifold subspace modeling module, and ``$\mathrm{S.Reg.}$'' denotes the subspace regularization term. \changed{``FRM'' and ``FFM'' denote the Feature Rectification Module and Feature Fusion Module~\cite{cmnext,cmx}, respectively.} The `\textuparrow'' indicates that a higher value is preferred.}
\footnotesize
\begin{tabular*}{0.96\textwidth}{@{\extracolsep{\fill}}l cccc c cc}
\toprule
\textbf{Method} 
& \hspace{0.3em} \textbf{Angular Marker} \hspace{0.3em}
& \hspace{0.3em} \textbf{Mani.}   \hspace{0.3em}
& \hspace{0.3em}  \textbf{S.Reg.}  \hspace{0.3em}
& \hspace{0.3em}  \textbf{FRM}        \hspace{0.3em}
& \hspace{0.3em}  \textbf{FFM}        \hspace{0.3em}
& mIoU $\uparrow$
& mAcc $\uparrow$ \\
\midrule 
LFX   & $\checkmark$ & $\checkmark$ & $\checkmark$ & $\checkmark$ & $\checkmark$ & 76.45   & 83.49\\
Exp.1 & $\times$     & $\checkmark$ & $\checkmark$ & $\checkmark$ & $\checkmark$ & 75.21(-1.24) & 84.18(+0.69)\\
Exp.2 & $\times$     & $\checkmark$ & $\times$     & $\checkmark$ & $\checkmark$ & 75.02(-1.43) & 83.36(-0.13)\\
Exp.3 & $\times$     & $\times$     & $\times$     & $\checkmark$ & $\checkmark$ & 74.13(-2.32) & 82.71(-0.78)\\
Exp.4 & $\times$     & $\times$     & $\times$     & $\times$     & $\checkmark$ & 73.34(-3.11) & 82.37(-1.12)\\
Exp.5 & $\times$     & $\times$     & $\times$     & $\times$     & $\times$     & 71.89(-4.56) & 81.27(-2.22)\\
\bottomrule
\end{tabular*}
\label{tab:ablation_components}
\vspace{-1em}
\end{table*}

\begin{table*}[ht]
\centering
\caption{
\changed{Ablation on auxiliary observations. ``A$\mathrm{ux}$. 33*'' indicates that
the auxiliary observations are set to $33$ for S.Aperture images and $12$ for focal
stacks.} The `\textuparrow'' indicates that a higher value is preferred, and
\textdownarrow'' denotes that a lower value is better for the given evaluation
metric. ``W.O'' denotes without FoP-ASM, and ``W.'' denotes with FoP-ASM.
\changed{``GFLOPs'' denotes the computational cost.}
\changed{For each metric column, the parenthetical value reports the signed
change relative to the corresponding result under Aux.2.}
}
\label{ab:b}
\footnotesize
\renewcommand{\arraystretch}{0.85}
\setlength{\tabcolsep}{6pt}
\begin{tabular*}{0.96\textwidth}{@{\extracolsep{\fill}}lcccccc}
\toprule
\multirow{2}{*}{\textbf{Views}} 
& \multicolumn{3}{c}{\textbf{Segmentation}}
& \multicolumn{3}{c}{\textbf{Salient Detection (FS)}} \\
\cmidrule(lr){2-4} \cmidrule(lr){5-7}
& mIoU $\uparrow$ & mAcc $\uparrow$ & \changed{GFLOPs}
& MAE (W.O) $\downarrow$ & MAE (W.) $\downarrow$ & \changed{GFLOPs} \\
\midrule
Aux.2   & 75.07 & 83.92       & \changed{63.06}  & 0.037 & 0.034       & \changed{15.05} \\
Aux.5   & 76.45(+1.38) & 83.49(-0.43) & \changed{67.59}  & 0.039(+0.002) & 0.032(-0.002) & \changed{16.13} \\
Aux.9   & 76.03(+0.96) & 83.82(-0.10) & \changed{73.62}  & 0.035(-0.002) & 0.032(-0.002) & \changed{17.57} \\
Aux.33* & 77.34(+2.27) & 85.50(+1.58) & \changed{109.82} & 0.036(-0.001) & 0.030(-0.004) & \changed{18.65} \\
\bottomrule
\end{tabular*}
\vspace{-1em}
\end{table*}

\begin{figure*}[ht]
\centering
\includegraphics[width=0.86\textwidth]{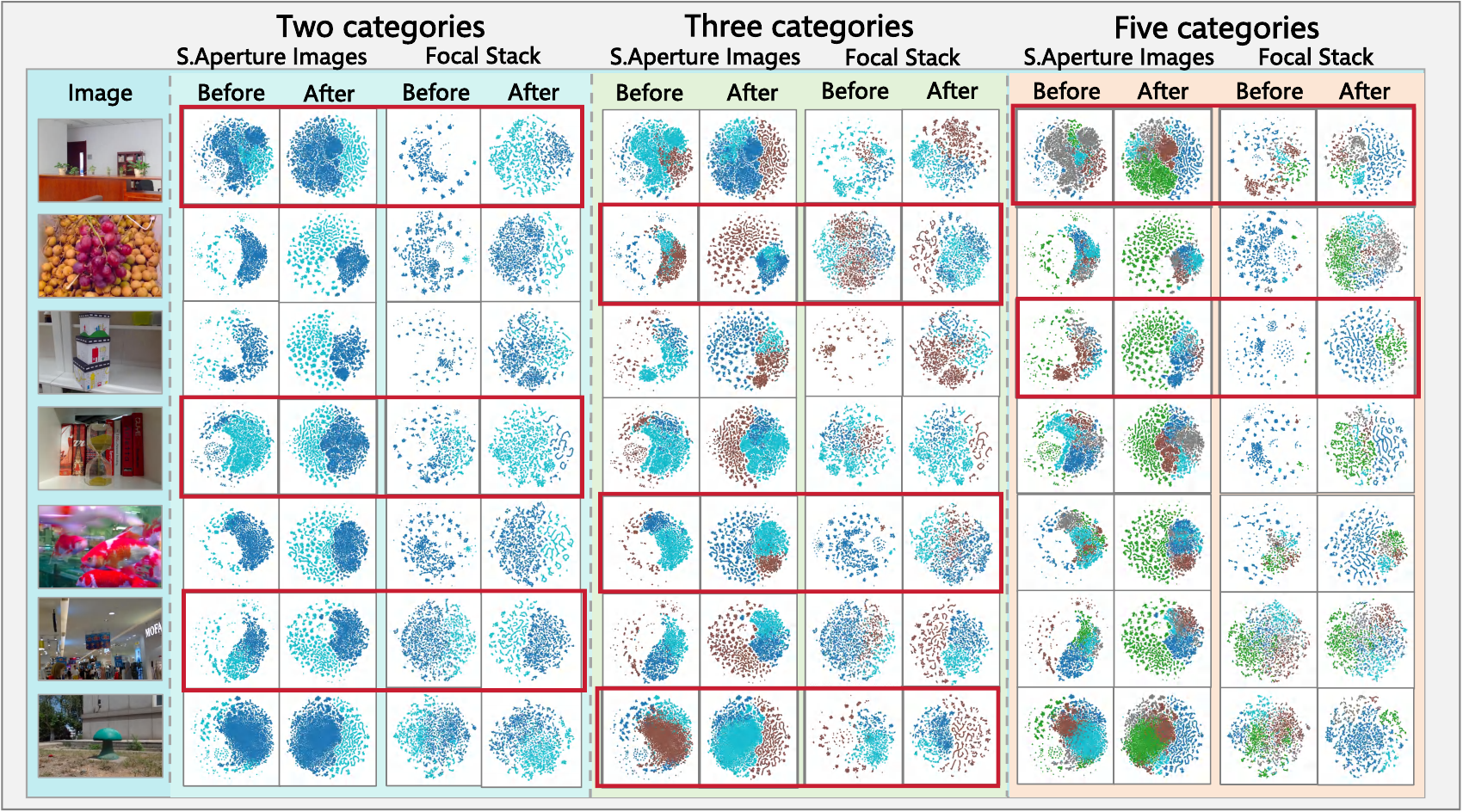}
\caption{
t-SNE visualization of central image features before and after the FoP-ASM module \changed{on the DUTLF-V2 dataset~\cite{DUTLFv2}}. Each row corresponds to one image sample.
To analyze feature organization under different semantic granularities, the t-SNE clustering is performed with 2, 3, and 5 categories.
\changed{For each setting, we compare two representative LF forms: S.Aperture images (disparity-centric form) and focal stack (refocus-centric form)}. 
}
\label{fig:lfx_tsne_visualization}
\vspace{-1em}
\end{figure*}

\subsection{Ablation for Core Components} \label{5_1}
As shown in Tab.~\ref{tab:ablation_components}, the number of mIoU consistently decreases as components are progressively removed from the full LFX model. 
\changed{Removing the angular marker} results in a $1.24\%$ mIoU drop, highlighting its role in encoding cross-view angular cues. Further removing the manifold subspace modeling and subspace regularization leads to additional degradation, indicating its effectiveness in constraining angular residuals within a shared subspace. \changed{For completeness, we additionally report an ablation in which the Feature Rectification Module (FRM) and Feature Fusion Module (FFM), two key components inherited from our baseline~\cite{cmnext,cmx}, are removed. FRM rectifies features from the two input branches, whereas FFM fuses the rectified features for subsequent prediction. Removing both modules reduces the mIoU to $71.89\%$, demonstrating the contribution of feature rectification and fusion to effective representation learning.} Furthermore, to verify whether FoP-ASM brings consistent representation enhancement across different LF forms, the central image features before and after applying the FoP-ASM module are visualized. 
As shown in Fig.~\ref{fig:lfx_tsne_visualization}, the figure is conducted on the central image under three settings, i.e., two, three, and five classes. 
After FoP-ASM is introduced, the feature distributions exhibit more compact intra-class structures and clearer inter-class separation. 
\changed{This trend is observed on both S.Aperture images and FS forms, demonstrating that LFX can effectively adapt to heterogeneous LF forms while improving feature discriminability.}

\begin{table*}[!htbp]
\centering
\caption{Stage-wise ablation for FoP-ASM. We report the performance, number of parameters, and computational overhead when progressively removing FoP-ASM from different encoder stages. The symbols        ``$\checkmark$'' and ``$\times$'' indicate whether FoP-ASM is applied to each encoder stage, ranging from Stage 1 to Stage 4. Those experiments progressively remove FoP-ASM from later encoder stages. ``P$\mathrm{arams}$.'' is the abbreviation for parameters. ``E$\mathrm{xp}$.'' is the abbreviation for experiment.}
\footnotesize
\begin{tabular*}{0.96\textwidth}{@{\extracolsep{\fill}}l cccc cc cc}
\midrule
\textbf{Method} 
& \hspace{0.3em} \textbf{Stage 1} \hspace{0.3em}
& \hspace{0.2em} \textbf{Stage 2} \hspace{0.2em}
& \hspace{0.2em} \textbf{Stage 3}\hspace{0.2em}
& \hspace{0.2em} \textbf{Stage 4} \hspace{0.2em}
& mIoU $\uparrow$
& mAcc  $\uparrow$
& Params. & GFLOPs\\
\midrule 
LFX   & $\checkmark$ & $\checkmark$ & $\checkmark$ & $\checkmark$  & 76.45   & 83.49 & 58.99 $M$ & 67.59 $G$\\
Exp.1 & $\checkmark$ & $\checkmark$ & $\checkmark$ & $\times$     & 75.90(-0.55) & 83.40(-0.09) & -0.16 $M$ & -0.21 $G$  \\
Exp.2 & $\checkmark$ & $\checkmark$ & $\times$     & $\times$     & 75.21(-1.24) & 82.27(-1.22) & -66.06 $K$ & -0.33 $G$\\
Exp.3 & $\checkmark$ & $\times$     & $\times$     & $\times$     & 74.96(-1.49) & 82.43(-1.06) & -11.10 $K$ & -0.22 $G$\\
Exp.4 & $\times$     & $\times$     & $\times$     & $\times$     & 74.13(-2.32) & 82.71(-0.78) & -3.52 $K$ & -0.28 $G$ \\

\midrule
Total:
& \multicolumn{8}{c}{Performance $74.13$ $\rightarrow$ $76.45$}
 Parameters {$58.74$$ M$ $\rightarrow$ $58.99$ $M$} \quad\quad Computation {$66.55$ $G$ $\rightarrow$ $67.59$ $G$}   \\  \bottomrule

\end{tabular*}
\label{ab:4}
\vspace{-1em}
\end{table*}

\subsection{Ablation on auxiliary observations} \label{5_2}
As shown in Tab.~\ref{ab:b}, we further investigate the impact of varying numbers of auxiliary observations on model performance.
Specifically, for semantic segmentation, the performance of LFX improves when increasing the \changed{number of S.Aperture images} from $\text{Aux}.2$ to $\text{Aux}.5$, where the mIoU rises.
When further increasing the views to $\text{Aux}.9$, the performance remains comparable. The slight performance fluctuation indicates that additional views may introduce inconsistent information. Nevertheless, when a larger number of observations is incorporated (\ie, $\text{Aux}.33$), the performance further improves, reaching $77.34\%$ mIoU and $85.50\%$ mAcc, demonstrating that sufficiently rich angular observations provide stronger cross-view geometric cues. For salient object detection, the results show that incorporating FoP-ASM reduces the MAE across different view settings. Under the $\text{Aux}.2$ configuration, the MAE decreases from $0.037$ to $0.034$, and continues to decrease as more views are introduced, reaching $0.030$ under $\text{Aux}.33$.
\changed{Meanwhile, the computational cost is reported. As the number of input observations
increases, the computational cost rises, while the predictive performance
exhibits an overall improving trend.}
These results demonstrate that FoP-ASM effectively leverages angular information while remaining compatible with different view configurations.

\begin{table*}[!htbp]
\centering
\caption{\changed{Ablation study on the construction of the angular marker.} ``Differ.Map'' denotes the difference map between each auxiliary observation and the central image. ``Ch.P'' indicates $P_s$, while ``So.P'' denotes $A_s$. ``Add.'' and ``Cat.'' denote element-wise addition and concatenation, respectively.}
\label{tab:ab_marker}
\footnotesize
\begin{tabular}{lcccccc}
\toprule
\textbf{Method} & \hspace{0.6em} \textbf{Differ. Map} \hspace{0.6em}
& \hspace{0.6em} \textbf{Ch.P} \hspace{0.6em} & \hspace{0.6em} \textbf{So.P} \hspace{0.6em}
& \hspace{0.6em} \textbf{ADD.} \hspace{0.6em} & mIoU $\uparrow$ & mAcc $\uparrow$ \\
\midrule
LFX   & $\checkmark$ & $\checkmark$ & $\checkmark$ & Cat. & 76.45 & 83.49 \\
Exp.1 & $\checkmark$ & $\checkmark$ & $\checkmark$ & $\checkmark$        & 76.10(-0.35) & 83.71(+0.22) \\
Exp.2 & $\checkmark$ & $\times$     & $\checkmark$ & $\times$     & 75.50(-0.95) & 83.02(-0.47) \\
Exp.3 & $\checkmark$ & $\checkmark$ & $\times$     & $\times$     & 75.79(-0.66) & 82.74(-0.75) \\
Exp.4 & $\times$     & $\times$     & $\times$     & $\times$     & 75.21(-1.24) & 84.18(+0.69) \\
\bottomrule
\end{tabular}
\vspace{-1em}
\end{table*}

\subsection{Stage-wise Ablation for FoP-ASM} \label{5_3} 
Tab.~\ref{ab:4} evaluates the effect of inserting FoP-ASM into different backbone stages. Overall, the mIoU improves steadily as FoP-ASM is applied to more stages. Without FoP-ASM, the model achieves $74.13\%$ mIoU and $82.71\%$ mAcc. Introducing FoP-ASM only at Stage~1 increases the mIoU to $74.96\%$, while extending it to Stages~$1$ to $2$ further improves the result to $75.21\%$. Finally, inserting FoP-ASM into all four stages achieves the best performance, reaching $76.45\%$ mIoU and $83.49\%$ mAcc. These results suggest that multi-stage angular modulation progressively enhances cross-view feature learning across the hierarchy. The mAcc shows mild fluctuations. This is expected since mAcc is more sensitive to class-frequency bias, while FoP-ASM mainly improves the overlap quality of predicted regions reflected by mIoU.

\begin{figure}[!t]
    \centering
    \includegraphics[width=0.95\columnwidth]{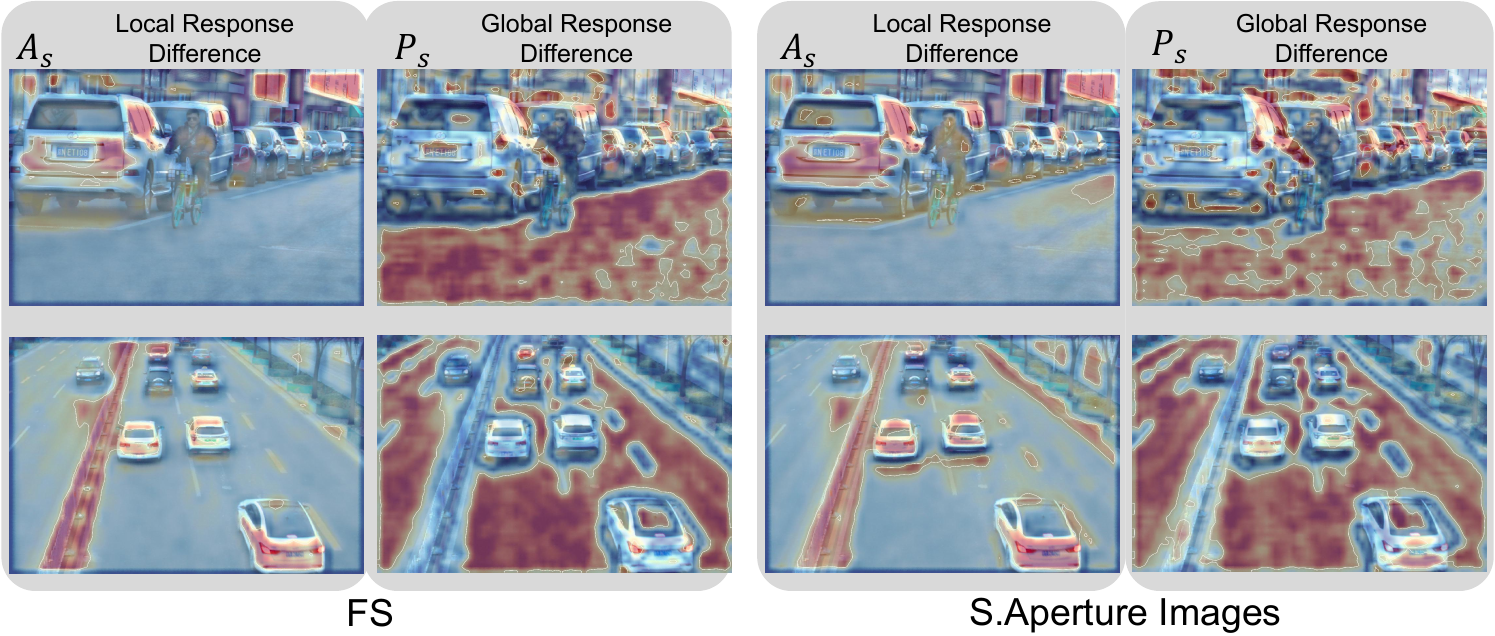}
    \caption{Heatmap visualization of \changedv{global response difference} and \changedv{local response difference} for S.Aperture images and FS.}
    \label{fig:angular-response-heatmaps}
    \vspace{-1em}
\end{figure}

\begin{table}[ht]
    \centering
    \caption{Performance and efficiency comparison on UrbanLF-real under the
    FS form. ``A$\mathrm{vg.R}$'' denotes the average ranking.}
    \label{tab:r1c4-fs-efficiency}
    \scriptsize
    \renewcommand{\arraystretch}{1.08}
    \setlength{\tabcolsep}{1.2pt}
    \begin{tabular*}{\columnwidth}
    {@{\extracolsep{\fill}}lcccccc@{}}
        \toprule
        Method
        & Acc $\uparrow$
        & mAcc $\uparrow$
        & mIoU $\uparrow$
        & Avg.R $\downarrow$
        & Params. $\downarrow$
        & GFLOPs $\downarrow$ \\
        \midrule
        SHIFNet$_5$~\cite{zhao2025unveiling}
        & 91.31 & 84.51 & 76.01 & 4.33 & 78.47M & 416.79 \\
        SACNet$_5$~\cite{wang2024alignment}
        & 91.94 & 85.58 & 77.93 & 2.33 & 327.66M & 596.85 \\
        CMNeXt$_5$~\cite{cmnext}
        & 91.99 & 84.83 & 77.16 & 2.67 & 116.61M & 128.51 \\
        StitchFusion$_5$~\cite{stitchfusion}
        & 91.49 & 83.37 & 75.32 & 4.67 & 69.98M & 310.41 \\
        \midrule
        \textbf{LFX$_5$ (Ours)}
        & \textbf{92.65}
        & \textbf{86.46}
        & \textbf{79.89}
        & \textbf{1.00}
        & 116.86M
        & 129.59 \\
        \bottomrule
    \end{tabular*}
    \vspace{-0.5em}
\end{table}

\subsection{\changed{Ablation of the Angular Marker}}  \label{5_4}
    As shown in Tab.~\ref{tab:ab_marker}, when concatenation is replaced with element-wise addition, the mIoU slightly decreases to $76.10\%$. Further removing the $P_s$ leads to a more noticeable drop to $75.50\%$. When the $A_s$ is removed instead, the performance decreases to $75.79\%$, which indicates the complementary contribution. Finally, \changed{removing the angular marker entirely causes} the largest degradation, reducing the mIoU to $75.21\%$. This result highlights the importance of angular cues for auxiliary observations. \changed{Meanwhile, as shown in Fig.~\ref{fig:angular-response-heatmaps}, we visualize
    the heatmaps of \changedv{global response difference} and \changedv{local response difference}. \changedv{Global response difference} primarily exhibits a global response, representing feature
    differences relative to the central image over large, smooth regions.
    \changedv{Local response difference} exhibits local response differences,
    particularly around object boundaries, occlusion edges, and fine-texture
    regions.}

\subsection{Efficiency Comparison with existing methods} \label{5_5}
 We compared LFX with recent methods under 
the UrbanLF-real FS setting, as reported in Table~\ref{tab:r1c4-fs-efficiency}. LFX 
achieves the best predictive performance across all three metrics with $116.86$M parameters 
and $129.59$ GFLOPs. Its computational cost is comparable to that of CMNeXt~\cite{cmnext}, introducing only 
$0.25$M additional parameters and $1.08$ additional GFLOPs while improving mIoU by $2.73$ 
percentage points. Compared with SACNet~\cite{wang2024alignment}, LFX achieves better performance with substantially 
fewer parameters and lower computational cost. Although LFX uses more parameters than SHIFNet~\cite{zhao2025unveiling} 
and StitchFusion~\cite{stitchfusion}, it requires substantially fewer GFLOPs than both methods while delivering 
clearly better performance. Overall, LFX achieves a favorable balance among predictive performance, 
parameter count, and computational cost.

\FloatBarrier
\section{Conclusion}
In this work, we present LFX, \changed{a form-flexible framework for light field perception that accommodates heterogeneous LF forms.} By introducing FoP-ASM, which assigns observation-specific angular markers and enforces shared manifold subspace constraints, LFX models observation differences and global semantic consistency without relying on form-specific geometric assumptions.

While LFX demonstrates strong adaptability across heterogeneous LF forms and perception tasks, the current framework mainly focuses on form-level modulation with controlled lighting conditions. Extending the framework to more diverse lighting conditions and various scenarios could further enhance its practical applicability. 
Exploring such directions will be an important avenue for future research. \changed{Furthermore, investigating
how to exploit LF formation characteristics within a joint-training paradigm
to develop an angular-aware LF foundation model represents another promising
direction.}

\FloatBarrier

\bibliographystyle{IEEEtran}
\bibliography{main}

\end{document}